\pdfoutput=1

\documentclass[11pt]{article}

\usepackage{eacl2023}

\usepackage{times}
\usepackage{latexsym}
\usepackage[T1]{fontenc}
\usepackage[utf8]{inputenc}
\usepackage{microtype}
\usepackage{inconsolata}
\usepackage{graphicx}
\usepackage{enumitem}

\newcommand{\commentout}[1]{}

\newcommand{\modelname}[1]{\textit{#1}} 
\newcommand{\rolename}[1]{\textsc{#1}}
\usepackage{enumitem}

\usepackage{todonotes}

\title{Multi-Task Learning for Joint Semantic Role and Proto-Role Labeling}

\author{Aashish Arora \and Harshitha Malireddi \and Daniel Bauer \\
Columbia University \\
New York, NY, USA\\
\texttt{\{aa4830,vm2662,db2711\}@columbia.edu}
\AND
Asad Sayeed\\
Dept. of Philosophy, Linguistics, \\and Theory of Science\\
University of Gothenburg, Sweden\\
  \texttt{asad.sayeed@gu.se} \\
 \And
  Yuval Marton \\
  University of Washington \\
  WA, USA \\
  \texttt{ymarton@uw.edu} \\}

\begin{document}
\maketitle
\begin{abstract}
We put forward an end-to-end  multi-step machine learning model which jointly labels semantic roles and the proto-roles of \citet{dowty1991}, given a sentence and the predicates therein. Our best architecture first learns argument spans followed by learning the argument's syntactic heads. This information is shared with the next steps for predicting the semantic roles and proto-roles. We also experiment with transfer learning from argument and head prediction to role and proto-role labeling. We compare using static and contextual embeddings for words, arguments, and sentences. Unlike previous work, our model does not require pre-training or fine-tuning on additional tasks, beyond using off-the-shelf (static or contextual) embeddings and supervision. 
It also does not require argument spans, their semantic roles,  and/or their gold syntactic heads as additional input, because it learns to predict all these during training.  
Our multi-task learning model raises the state-of-the-art predictions for most proto-roles.

\end{abstract}

\section{Introduction}\label{sec:intro}

We jointly learn semantic role labelling (SRL) and semantic proto-role labelling (SPRL), demonstrating that multi-task learning benefits both tasks in English.
Current work on SRL does not make use of proto-roles \citep{dowty1991} at all, while existing work on SPRL \citep{teichert2017, rudinger2018,opitz-frank-2019-argument} focuses on predicting proto-role properties themselves, and views SRL merely as an external task for 
pre-training
or additional input.

In this work, we test a hypothesis that joint learning should yield better results by sharing predicate, argument, and sentence representations between the SRL and SPRL tasks. Instead of only relying on gold arguments' spans and roles, like \citet{rudinger2018}, we optionally train an explicit step in our pipeline to predict ``nameless'' (role-unaware) argument spans and their syntactic heads.  We experiment with various input representations, using a single embedding for each argument head (like \citet{rudinger2018}), using embeddings for entire argument spans, and using also entire sentence embeddings.  
We experiment with both static  (GloVe) and contextual  (BERT) embeddings.

Frame Semantics~\citep{fillmore1968} and SRL has proven useful for extracting semantic information of facts and events from text (roughly who did what to whom, where, when and/or how), quickly and on scale. 
Traditionally, SRL systems are trained either on PropBank  or FrameNet annotations (\S\ref{sec:relwork}). 
Proto-roles \citep{dowty1991,reisinger2015} break down these more traditional semantic roles to sub-components (proto-role properties), such as animacy, sentience, change of state, etc. This has potential to improve model learnability (feature-sharing, less sparse features).

The prior work on SPRL (\citet{rudinger2018}, hereafter RvD) 
relies on gold argument spans and gold syntax (to identify argument heads). In addition, RvD uses a variety of auxiliary tasks for pre-training, including neural MT, requiring additional annotated data. This makes that approach less applicable to many languages, harder to replicate, and more cumbersome and slow to use (especially on large input sizes). In our work, we do not fine-tune on these auxiliary tasks. 
\citet{opitz-frank-2019-argument} incorporate BERT embeddings for arguments and leave out gold syntax, but their work is still dependent on the identification of gold spans. We avoid dependence on gold spans.

Our contributions are:
\setlist{nolistsep}
\begin{itemize}[noitemsep]
    \item A new joint multi-task learning (MTL) multi-step model for SRL and SPRL.
    \item We 
    set new state of the art
    in SPRL prediction quality (micro-averaged F1)
    without requiring pretraining on auxilary tasks (other than for GloVe and BERT), unlike prior work. 
    \item Compared to the stand-alone SRL and SPRL baselines, our joint models show  gains in prediction quality for both SRL and proto-role property prediction.
    \item Our approach avoids requiring syntax or gold argument spans in the input. Instead it works with predicted (nameless) argument spans and head words. Embedded span and head word representations are shared between the SRL and SPRL predictors.
    We also explore transfer learning from the span/head prediction steps to the SRL/SPRL prediction steps, and demonstrate that this improves SPRL. 
   \item We experiment with both static (GloVe) or contextual embeddings (BERT): adding contextualized and full sentence embeddings further improves SPRL performance. 
\end{itemize}

\begin{figure*}[ht]
    \centering
    \includegraphics[width=\textwidth]{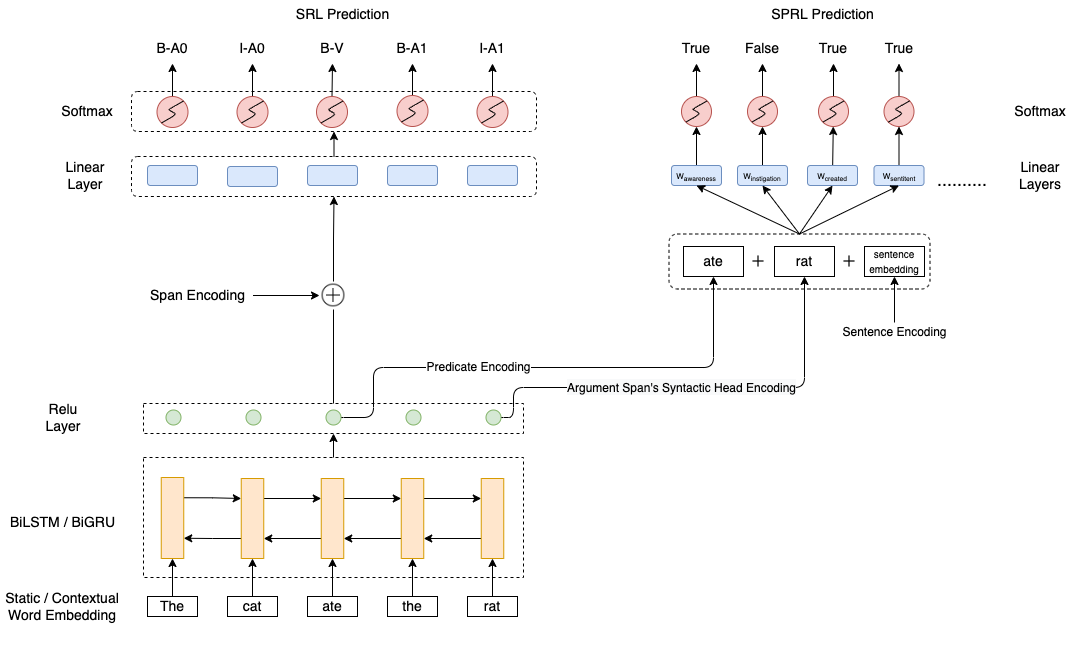}
    \caption{Multi-task learning model for jointly predicting semantic roles and proto-roles.}
    \label{fig:srl-sprl}
\end{figure*}
\begin{figure*}
    \centering
    \includegraphics[width=\textwidth]{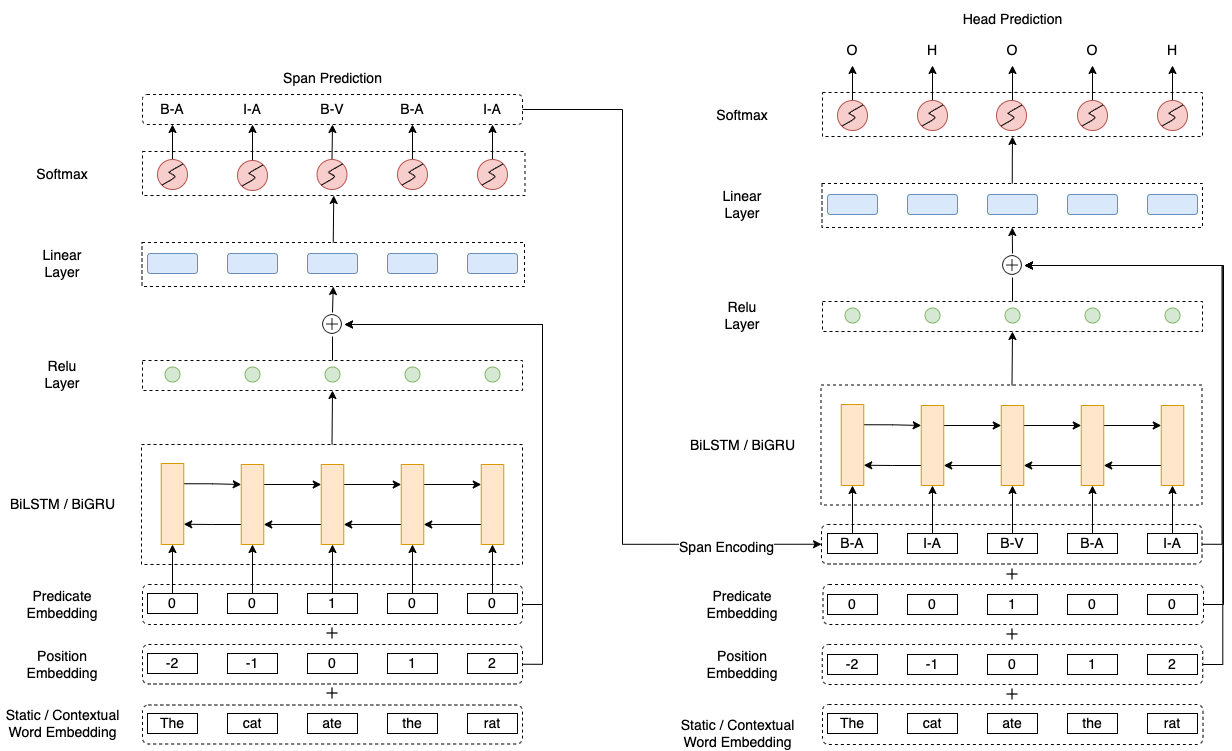}
    \caption{Sequential model for Span and Head prediction}
    \label{fig:span-head}
\end{figure*}

\section{Related Work}
\label{sec:relwork}
Semantic representations of natural language typically focus on the relations between a predicate and its semantic arguments \cite{abend2017}. Semantic roles are types of argument relations, such as Agent, Patient, or Theme \cite{fillmore1968}. 
It
is challenging to agree on an inventory of roles that captures all possible argument types across all  predicates. For example, should there be a Beneficiary or Recipient role, distinct from the Patient role?  
Frame semantics \cite{fillmore2003} groups predicates into semantic frames. 
Each frame has its own specific inventory of semantic roles.
\citet{dowty1991}, in contrast, suggests that a discrete set of semantic roles is unnecessary to explain how semantic arguments are realized syntactically. Instead, he characterizes arguments by bundles of properties they may have. 
Dowty identifies two clusters of such properties, an \rolename{Agent} proto-role and a \rolename{Patient} proto-role. For instance, an argument similar to the \rolename{proto-agent} might have the properties of being sentient and having volitional involvement in the event.
\footnote{Following related work, we use the term ``proto-role'' to refer to individual properties, such as \textit{awareness}, rather than the \rolename{proto-agent} and \rolename{proto-patient} prototypes.} For a full-list of properties used in this paper, refer to Table~\ref{tab:sprl-static-emb}.

Most work on SRL has used PropBank \cite{palmer2005}, which
uses a limited set of role labels \rolename{Arg0},$\ldots$, \rolename{Arg5}.\footnote{We ignore modifiers and other role types here.}
In general, \rolename{Arg0} corresponds to the \rolename{proto-agent} and \rolename{Arg1} to the \rolename{proto-patient}. The remaining numbered argument labels are interpreted on a per-predicate basis (with some cross-predicate regularities).
Work on proto-role annotation, which involves annotating or predicting bundles of Dowty-inspired proto-role properties for each argument, is more recent \cite{reisinger2015}. Much of it has been undertaken within the framework (and dataset) of Universal Decompositional Semantics (UDS; \citealp{white2016,white2020}). UDS annotates predicate-argument structures from the Universal Dependency data \cite{deMarnefee2021} with fine-grained semantic properties for both predicates and arguments, including proto-role properties for arguments. 

The availability of datasets with proto-role annotations \cite{reisinger2015, white2016, white2020} has resulted in an interest in predicting property bundles automatically. \citet{teichert2017} model the problem using a conditional random field (CRF), in which each predicate/argument/property combination is a binary variable whose value needs to be inferred. \citet{rudinger2018} (RvD) predict each proto-role property independently for each argument span. Their approach  first encodes the input sentence using a BiLSTM, then combines the latent representation for the predicate and the head word of each argument span into a single ``neural-Davidsonian'' representation (after 
\citealp{davidson:1967}). 
Parameters for their model are additionally optimized on a variety of external auxiliary tasks, such as English$\rightarrow$French neural machine translation, super-sense tagging, and PropBank-based SRL. 
\citet{opitz-frank-2019-argument} use a BiLSTM training process similar to RvD, using gold SRL spans and static GLoVe embeddings, and additionally add BERT embeddings to their model. They train 50 different models with random initialization and use them in an ensemble setup to produce 18 weight matrices, one for each property.

Multi-task learning has previously been employed in SRL. \citet{cai2019} train an LSTM based SRL model with dependency-parsing as a joint auxiliary task.
\citet{conia2021} 
train
joint cross-lingual / cross-representational sentence and predicate/argument encoders and then use a set of language specific decoders to predict SRL annotations for each language and semantic framework. 

\commentout{
Another difference is that we approach SRL as a BIO (beginning/in/out) tagging problem \cite{collobert2011, he-etal-2017-deep, strubell-etal-2018-linguistically}, while \citeposs{rudinger2018} auxiliary SRL task predicts roles directly,
given the predicate and (gold) argument head word as input (more similar to \citet{he-etal-2018-jointly}).
}

\section{Methodology}
\label{sec:method}

\subsection{Model Architecture}
Our overall model pipeline is described in Figure~\ref{fig:srl-sprl}. 
Assuming that feature/weight-sharing is likely to yield better results than a cascaded setting (using SRL output as
input to SPRL as in RvD),
our MTL architecture uses a single BiLSTM layer shared between upstream SRL and SPRL tasks. In the MTL setting, the parameters of this layer are optimized during training on a combined loss for all tasks: the multinomial SRL sequence tagging, and the~18 proto-role binary classifiers.  The output of the Bi-LSTM layer is passed through an additional ReLU layer 
before final predictions.

 We start with an embedding for each token, either static (GloVe; \citet{pennington-etal-2014-glove}) or contextualized on the sentence (BERT; \citet{devlin-etal-2019-bert}). 
We concatenate the embeddings with a predicate indicator (1 if the token is a predicate, 0 otherwise) and position information relative to the predicate \footnote{Note that we use gold labels for predicates}. For instance, the position information for the  token `rat' in the sentence, ``The cat ate the rat'' would be 5-3=2. We use a BiLSTM to predict ``nameless'' argument spans (i.e., spans without role names yet). For each token, we predict if it is the start of an argument span (B-A), is inside an argument span (I-A), is a verb (B-V) or none of these (O).  Then, this argument encoding is passed as an additional input to the head prediction component, which predicts if each token is a head word (H) or not (O). 
The resulting span and head information is then shared between the SRL and SPRL prediction components. 
This allows us to have a fully end-to-end model for joint SRL and SPRL, in the sense that no gold span or syntax information is required as input.
The argument span and head prediction module is described in Figure \ref{fig:span-head}. 
Note that typically non-syntax-based SRL tasks does not use head information, but here instead of using a syntactic parser, we simply predict each argument's head (an easier task then full syntactic parsing) and test if this additional explicit representation helps.

The SPRL component is essentially the same as in 
RvD.
We represent each predicate/argument pair by concatenating the encoded predicate and encoded argument head (from the BiLSTM encoder). This representation is then passed to 18 separate binary classifiers, each having its own linear transformation and a softmax layer. 
With GloVe, we additionally experiment with adding complete span embeddings to each predicate/argument pair representation, by averaging in-span embeddings\footnote{excluding stopwords listed in NLTK \cite{bird-loper-2004-nltk}, but including prepositions and pronouns}.
We also optionally concatenate complete sentence embeddings to the pair representation, again computed by averaging token embeddings. With BERT we use the special CLS embedding as the sentence representation (\S\ref{sec:exp-contextual}).

The SRL component  predicts a semantic role tag (such as B-ARG0, I-ARG0, or O) for each token in the sentence, making the nameless spans ``named'' with role tags. The input to this component is similar to the input for the head predictor. For each token, we concatenate the encoded token representation, predicate indicator, and positional information. We optionally concatenate span and/or head information to the input (either gold or predicted). We pass this representation through a linear layer and then a softmax layer.

\subsection{Experimental Setup}
\label{sec:exp-description}
\paragraph{Data}
We run our experiments on the proto-role data by \citet{reisinger2015}\footnote{The data is available at \url{http://decomp.io/projects/semantic-proto-roles/}} (called SPR1 by RvD
-- we use their training/dev/test split), which is based on the Wall Street Journal portion of the Treebank \cite{marcus-etal-1993-building}. SRL annotations from PropBank \citep{palmer2005} are available for most of this data. We filter out a small number of sentences for which we could not locate PropBank annotations. The resulting data includes 3615 sentences in the training set, 485 sentences in the dev set, and 511 sentences in the test set.
This rather small dataset size poses a challenge for some of our experiments, as we discuss in Section~\ref{sec:exp-contextual}.
We do not include the more recent UDS protorole annotations by \citet{white2016}, which are based on the English Web Treebank section of the Universal Dependencies data \citep{deMarnefee2021}, because gold PropBank annotations are not readily available for this data.
The data contains human-rated proto-role scores on a $1 \ldots 5$ scale. We convert the ratings to binary labels using a threshold of $>2$. Note that \citep{rudinger2018} use a threshold of $>3$, but we found that this results in few positive labels for some proto-roles on our limited data.

\paragraph{Metrics}
For SRL, we output argument-specific BIO tags for each token given a predicate. For SPRL, we output a binary proto-role property for each argument's syntactic head given a predicate. For each prediction, we report 
F1-score
and micro-averaged F1 overall.\footnote{Note that for SRL, the scores are computed per-token. Span-based metrics using partial overlap criteria rather than tag/token-based metrics have been more common for SRL evaluation. But tag/token-based metrics tell us how well we recover the exact boundaries of semantic constituents  more precisely than the commonly used span overlap metrics.}.

\section{Experiments and Results}
\label{sec:exp}

\subsection{Experiments Overview}

In our experiments, we compare our joint SRL and SPRL prediction task to learning SRL alone and SPRL alone. The latter setting, when using gold syntax to identify head words, is closest to RvD,
which  we take as our main baseline for SPRL.

We also experiment with using gold spans (from PropBank) and/or heads%
\footnote{Gold heads are obtained by converting the Treebank annotations to predicate (verb)-rooted dependencies using the Stanford head percolation rules \cite{de-marneffe-etal-2006-generating} implemented in the CoreNLP toolkit \cite{manning-etal-2014-stanford}}
vs.\ predicted spans or heads.
Next, we try transfer learning: pre-initializing the model weights of the upstream tasks with our span and head predictors' weights, instead of only providing the raw embeddings as input.
 
Our initial set of experiments uses static GloVe embeddings \citep{pennington-etal-2014-glove} as input. In a second set of experiments, we try replacing these static embeddings with contextualized embeddings from BERT \citep{devlin-etal-2019-bert}. The initially low performance of this approach leads us to hypothesize that the model may have too many parameters, given the limited amount of training data. We then systematically optimize the hyperparameters of our model to reduce model complexity: we replace the BiLSTM layer with a BiGRU layer \citep{chung2014}, which has fewer trainable parameters, and set the hidden dimension of the GRU to 8 (after performing binary search between 2 and 768 dimensions). We also remove the linear layer after the ReLU activation function. Finally, we set the rate of the dropout layer to 25\% (after performing binary search between 20\% and 30\%).

All experiments use the Adam optimizer with a learning rate of  0.001 and a Categorical Cross-Entropy loss for both SRL and SPRL.

\begin{table*}[!t!hb]
\centering
\resizebox{\textwidth}{!}{%
\begin{tabular}{|l|r|r||r||r|r|r|r||r|}
\hline
\textbf{Semantic proto-role} & 
\begin{minipage}[t]{0.7in}%
\textbf{RvD Best Reported Results}%
\end{minipage}
 & 
\begin{minipage}[t]{0.4in}%
 \textbf{RvD SPR1 Basic Model}%
\end{minipage}
 & 
\begin{minipage}[t]{0.75in}%
\textbf{SPRL-only, Replicated Baseline, gold heads}%
\end{minipage} 
 & 
\begin{minipage}[t]{0.65in}%
\textbf{MTL Baseline, gold heads}%
\end{minipage}  
 & 
\begin{minipage}[t]{0.6in}%
\textbf{MTL, gold spans + \\heads}%
\end{minipage}
 & 
\begin{minipage}[t]{0.8in}%
\textbf{MTL, shared span wts, \\gold spans + heads + sent}%
\end{minipage} 
 & 
\begin{minipage}[t]{0.8in}%
\textbf{MTL, shared span + head wts, \\gold span + head + sent}%
\vspace{.05in}
\end{minipage}  
 & 
\begin{minipage}[t]{0.65in}%
\textbf{MTL, predicted span + \\head + \\sent}%
\end{minipage}  
\\
\hline
awareness&89.9&88.3&87.0&87.0&86.0&84.0&89.0&86.0\\
change\_of\_location&45.7&60.0&77.0&78.0&79.0&78.0&79.0&75.0\\
change\_of\_state&71.0&66.8&91.0&92.0&90.0&92.0&94.0&93.0\\
changes\_possession&58.0&57.1&32.0&33.0&34.0&19.0&16.0&17.0\\
existed\_after&85.9&86.9&84.0&89.0&89.0&88.0&88.0&87.0\\
existed\_before&85.1&86.0&87.0&85.0&88.0&85.0&87.0&86.0\\
existed\_during&95.0&94.2&93.0&94.0&95.0&94.0&95.0&95.0\\
exists\_as\_physical&82.7&82.3&80.0&77.0&80.0&77.0&78.0&76.0\\
instigation&88.6&84.6&85.0&86.0&84.0&86.0&85.0&85.0\\
location\_of\_event&53.8&46.9&23.0&30.0&38.0&33.0&34.0&32.0\\
makes\_physical\_contact&47.2&52.7&79.0&75.0&77.0&81.0&79.0&77.0\\
manipulated\_by\_another&86.8&82.2&82.0&82.0&84.0&73.0&84.0&82.0\\
pred\_changed\_arg&70.7&67.4&63.0&65.0&65.0&68.0&70.0&70.0\\
sentient&90.6&89.6&88.0&89.0&86.0&85.0&86.0&82.0\\
stationary&47.4&43.2&70.0&68.0&68.0&70.0&68.0&65.0\\
volition&88.1&87.9&86.0&83.0&85.0&85.0&86.0&83.0\\
created&39.7&46.6&24.0&42.0&26.0&30.0&38.0&38.0\\
destroyed&24.2&11.1&19.0&8.0&16.0&12.0&21.0&21.0\\ 
\hline
micro-F1&83.3&81.7&85.0&85.0&85.0&84.0&86.0&85.0\\
macro-F1&69.6&70.5&69.4&70.2&70.6&68.9&70.9&69.4\\
\hline
\end{tabular}}
\caption{F1-scores for SPRL tasks in both stand alone and MTL settings with static GloVe embeddings. RvD \citep{rudinger2018}'s results are not directly comparable with our results (see Replicated Baseline in Section \ref{sec:exp-static}).
}
\label{tab:sprl-static-emb}
\end{table*}

\begin{table*}[!t!hb]
\centering
\resizebox{\textwidth}{!}{%
\begin{tabular}{|l||r|r|r|r|r||r|r|r||r|r|}
\hline
\begin{minipage}[t]{0.4in}%
\textbf{Semantic Role}%
\end{minipage}
 & 
\begin{minipage}[t]{0.65in}%
\textbf{SRL-only Baseline}%
\end{minipage}
 & 
\begin{minipage}[t]{0.65in}%
 \textbf{SRL-only,  gold spans}%
\end{minipage} 
 &
\begin{minipage}[t]{0.65in}%
 \textbf{SRL-only,  gold spans + heads}%
\end{minipage}  
 & 
\begin{minipage}[t]{0.65in}%
 \textbf{SRL-only,  predicted spans}%
\end{minipage}  
 & 
\begin{minipage}[t]{0.65in}%
 \textbf{SRL-only,  predicted spans + heads}%
\end{minipage}  
 & 
 \begin{minipage}[t]{0.6in}%
 \textbf{MTL SRL - Baseline}%
\end{minipage}
 & 
\begin{minipage}[t]{0.5in}%
 \textbf{MTL, SRL gold spans}%
\end{minipage}  
 & 
\begin{minipage}[t]{0.75in}%
\textbf{MTL, SRL  gold spans + heads}%
\end{minipage} 
 &
\begin{minipage}[t]{0.75in}%
\textbf{MTL, SRL  predicted spans}%
\end{minipage} 
 & 
\begin{minipage}[t]{0.75in}%
\textbf{MTL, SRL  predicted spans + heads}%
\end{minipage}
 \\
\hline
B-V&100.0&100.0&100.0&100.0&100.0&100.0&99.0&100.0&100.0&100.0\\
B-A0&83.0&88.0&92.0&85.0&85.0&84.0&88.0&88.0&85.0&87.0\\
I-A0&78.0&88.0&94.0&83.0&84.0&78.0&89.0&88.0&83.0&84.0\\
B-A1&74.0&80.0&88.0&73.0&75.0&75.0&83.0&84.0&73.0&77.0\\
I-A1&63.0&81.0&88.0&63.0&65.0&65.0&85.0&83.0&65.0&64.0\\
B-A2&25.0&47.0&61.0&29.0&29.0&31.0&52.0&50.0&31.0&31.0\\
I-A2&23.0&45.0&53.0&21.0&25.0&24.0&48.0&38.0&29.0&32.0\\
B-A3&12.0&30.0&50.0&11.0&11.0&19.0&30.0&30.0&21.0&27.0\\
I-A3&0.0&3.0&26.0&6.0&6.0&7.0&12.0&8.0&6.0&6.0\\
B-A4&52.0&73.0&72.0&67.0&53.0&63.0&69.0&69.0&60.0&48.0\\
I-A4&34.0&38.0&58.0&45.0&34.0&41.0&46.0&45.0&48.0&35.0\\
B-A5&0.0&0.0&0.0&0.0&0.0&0.0&0.0&0.0&0.0&0.0\\
I-A5&0.0&0.0&0.0&0.0&0.0&0.0&0.0&0.0&0.0&0.0\\
O&88.0&99.0&100.0&89.0&89.0&89.0&100.0&100.0&89.0&89.0\\
\hline
micro-F1&80.0&91.0&94.0&80.0&81.0&81.0&93.0&92.0&81.0&82.0\\
macro-F1&45.1&55.1&63.0&48.0&46.9&48.3&57.2&55.9&49.3&48.6\\
\hline
\end{tabular}}
\caption{F1-scores for SRL tasks in both stand alone and multi-task learning (MTL) setting with static (GloVe) embeddings.
B- = beginning. B-V = beginning of the predicate. B-A0 for beginning of \rolename{Arg0} to B-A5 for beginning of \rolename{Arg5}. I- = in the argument's span. O = outside any predicate or argument span.
Support for B-A5 and I-A5 in our data is near 0, justifying the poor performance of those roles.
}
\label{tab:srl-static-emb}
\end{table*}

\subsection{SPRL-Only Baseline}
\label{sec:exp-static}
RvD's
best reported results (our Table~\ref{tab:sprl-static-emb}, column~2) cannot be directly compared to our results for several reasons: 
their models were pre-trained with external auxiliary tasks, such as neural MT, 
and their binarization of human-rated proto-role scores is different from ours (see above). Regardless, we include their reported results for context.

For a fairer comparison, we replicated RvD's \modelname{``SPR1 basic model"} using our threshold without any pre-training beyond using off-the-shelf GloVe embeddings (same as they did). 
The original \modelname{``SPR1 basic model"} results reported in
RvD
are repeated in our Table~\ref{tab:sprl-static-emb}, column~3,
and  our replication's results are in column~4 (\modelname{SPRL-only}).
Both receive gold predicate and argument head as input. Our results show similar trends as RvD. 
Scores for~10 of the~18 proto-roles are similar (1-2\% difference). However, scores for~5 other proto-roles 
are over~5\% higher than RvD's reported ones, and scores for~3 proto-roles are over~5\% lower.

\subsection{SRL-Only Baseline}
SRL results are reported in Table~\ref{tab:srl-static-emb}. 
Column~2 shows baseline results for SRL-only, without any span or head information in the input. 
Column~3 shows  adding span information improves average micro-F1 by 11\%. 
Column~4 
shows that additionally adding heads boosts performance by another 3\% on average, and improves prediction for 10 out of 14 tags, decreasing for only one tag. \modelname{SRL-only} results using predicted spans and heads (columns~5 and~6) were as expected worse than gold. Predicted heads also do not provide as much of an improvement over predicted spans only. Scores improved for only~5 roles and decreased for~2. The weighted F1-score slightly improved (from~80\% to~81\%).

\subsection{SRL+SPRL Multi-Task Learning}

We see that multi-task learning helps both 
SPRL (Table~\ref{tab:sprl-static-emb}, column~5 vs.\ column~4)
and
SRL (Table~\ref{tab:srl-static-emb}, column~7 vs.\ column~2),
compared to their respective stand-alone baselines, confirming our main hypothesis. 
Surprisingly, these gains for SPRL  show only in macro-F1 but not in micro-F1,
even though  we see F1 gains for~11 out of~18 proto-roles.
Specifically, we see large gains for \rolename{location\_of\_event}, \rolename{created} and \rolename{existed\_after}, but large drops for \rolename{destroyed} and \rolename{makes\_physical\_contact}. We have no explanation for this yet.

For SRL, we see gains for~10 out of~14 role tags, with  \texttt{B-A2, B-A3, B-A4} and \texttt{I-A4} performing significantly better. We also see a boost  of 1\% in micro-averaged F1, and over 3\% in macro-averaged F1. 
Additionally, we note that models in the multi-task learning setting converge faster, i.e. the MTL models require fewer training iterations (15 compared to~30 for our SPRL baseline). 

\paragraph{Gold span and head representations} 
Adding explicit gold information of spans and heads (Table~\ref{tab:sprl-static-emb}, column~6) slightly raises the macro-F1 to 70.6, but doesn't affect the micro-F1.

Pre-initializing upstream steps with the model weights of spans and heads ("shared wts" columns, aka transfer learning,
instead of only providing GloVe embeddings for input),
together with adding a fixed length sentence embedding to each predicate/argument pair representation (column~8), improves both micro and macro F1 scores to 86 and 70.9 respectively. We see pronounced gains for \rolename{destroyed} and \rolename{pred\_changed\_arg} but large drops for \rolename{changes\_possession}.
Curiously, sharing weights of spans but not heads (column~7) deteriorates the scores to 84 and 68.9 respectively.

\paragraph{Predicted spans and heads}

We took the most promising joint learning model for SRL+SPRL on gold input (Table~\ref{tab:sprl-static-emb}, column~8), and retrained it with predicted input (column~9).  
As expected, we notice a small dip in F1 score across all the SPRL properties. But it performs on the same level as that of the \modelname{MTL with gold span+head} (Table \ref{tab:sprl-static-emb}, column~5), suggesting that the addition of sentence embeddings and pre-initialization of  span/head prediction model weights is very useful.

For SRL
(Table~\ref{tab:srl-static-emb}, columns~5-6 for \modelname{SRL-only} and~10-11 for \modelname{MTL}), 
adding predicted head information hurts \texttt{B-A4} and \texttt{I-A4} prediction compared to passing only predicted spans, in both \modelname{SRL-only} and \modelname{MTL}  settings. The MTL model with predicted span+heads performs better than our SRL-only and MTL SRL baseline, showing gains for \texttt{B-A0, I-A0, B-A1, I-A2,} and \texttt{B-A3} semantic roles.   

\begin{table*}[!t!hb]
\centering
\resizebox{15cm}{!}{%
\begin{tabular}{|l||r|r||r|r||r|}
\hline
\textbf{Semantic proto-role} & 
\begin{minipage}[t]{0.8in}%
\textbf{SPRL-only,  biLSTM + \\gold head}%
\end{minipage}
 & 
\begin{minipage}[t]{0.8in}%
\textbf{SPRL-only,  LR + \\gold head}%
\end{minipage}
 & 
\begin{minipage}[t]{0.8in}%
\textbf{MTL,  GRU + gold head}%
\end{minipage}
 & 
\begin{minipage}[t]{1in}%
\textbf{MTL,  GRU + gold head + \\sentence emb.}%
\end{minipage}
 & 
\begin{minipage}[t]{1in}%
\textbf{MTL,  GRU + predicted head + sentence emb.}%
\vspace{0.05in}
\end{minipage}
  \\
\hline
awareness&80.0&74.0&79.0&89.3&83.0\\
change\_of\_location&0.0&29.0&67.0&82.6&78.0\\
change\_of\_state&49.0&49.0&61.0&69.3&62.0\\
changes\_possession&0.0&24.0&31.0&39.2&37.0\\
existed\_after&80.0&77.0&74.0&87.0&81.0\\
existed\_before&79.0&78.0&75.0&86.2&80.0\\
existed\_during&91.0&90.0&85.0&97.6&92.0\\
exists\_as\_physical&51.0&67.0&71.0&80.7&76.0\\
instigation&77.0&75.0&78.0&86.3&80.0\\
location\_of\_event&0.0&39.0&39.0&43.6&40.0\\
makes\_physical\_contact&0.0&40.0&71.0&82.2&78.0\\
manipulated\_by\_another&84.0&77.0&78.0&88.0&81.0\\
pred\_changed\_arg&47.0&49.0&65.0&73.5&69.0\\
sentient&56.0&71.0&76.0&88.7&84.0\\
stationary&0.0&29.0&65.0&72.2&68.0\\
volition&79.0&29.0&78.0&88.3&83.0\\
created&0.0&21.0&32.0&35.7&34.0\\
destroyed&0.0&20.0&15.0&23.9&22.0\\
\hline
micro-F1&52.8&71.5&77.0&87.3&83.9\\
macro-F1&42.9&52.1&63.3&73.0&68.2\\
\hline
\end{tabular}}
\caption{F1-scores for SPRL tasks in both stand alone and MTL setting with contextual embeddings (BERT).}
\label{tab:sprl-dyn-emb}
\end{table*}

\begin{table*}[!t!hb]
\centering
\resizebox{14cm}{!}{%
\begin{tabular}{|l||r|r|r||r|r||r|}
\hline
\begin{minipage}[t]{0.4in}%
\textbf{Semantic Role}%
\end{minipage}
 &
\begin{minipage}[t]{0.7in}%
\textbf{Baseline SRL-Only}%
\end{minipage}
 & 
\begin{minipage}[t]{0.7in}%
 \textbf{SRL-Only\\ Gold \\spans}%
\end{minipage} 
 & 
\begin{minipage}[t]{0.9in}%
 \textbf{SRL-Only\\ Gold\\ spans + heads}%
\end{minipage} 
 & 
\begin{minipage}[t]{0.7in}%
 \textbf{SRL-Only Predicted spans}%
\end{minipage} 
 & 
\begin{minipage}[t]{0.9in}%
 \textbf{SRL-Only Predicted spans + heads}%
\end{minipage} 
 & 
\begin{minipage}[t]{0.9in}%
 \textbf{MTL\\ Predicted spans + heads}%
 \vspace{.05in}
\end{minipage} 
\\
\hline
B-V & 100.0 & 100.0& 100.0 & 100.0 & 100.0 & 100.0   \\
B-A0 & 80.0 & 86.0 & 93.0 & 86.0 & 87.0 & 86.0   \\
I-A0 & 75.0 & 83.0 & 94.0 & 82.0 & 84.0 & 83.0 \\
B-A1 & 76.0 & 82.0 & 87.0 & 75.0 & 76.0 & 75.0 \\
I-A1 & 64.0 & 77.0 & 89.0 & 64.0 & 66.0 & 60.0  \\
B-A2 & 28.0 & 51.0 & 53.0 & 31.0 & 30.0 & 31.0  \\
I-A2 &  24.0 & 43.0 & 55.0 & 24.0 & 28.0 & 33.0 \\
B-A3 & 1.0 & 2.0 & 54.0 & 15.0 & 13.0 & 29.0  \\
I-A3 & 5.0 & 0.0 & 28.0 & 6.0 & 8.0 & 9.0  \\
B-A4 & 49.0 & 68.0 & 78.0 & 68.0 & 63.0 & 52.0  \\
I-A4 &  32.0 & 58.0 & 62.0 & 47.0 & 35.0 & 39.0 \\
B-A5 &  0.0 & 0.0 & 0.0 & 0.0 & 0.0 & 0.0  \\
I-A5 & 0.0 & 0.0 & 0.0 & 0.0 & 0.0 & 0.0 \\
O &  81.0 & 100.0 & 100.0 & 89.0 & 90.0 & 91.0 \\
\hline
micro-F1 & 75.2 & 90.6 & 94.4 & 80.86 & 80.0 & 81.8\\
macro-F1&43.9&53.6&63.8&49.1&48.6&49.1\\
\hline
\end{tabular}}
\caption{F1-scores for SRL tasks in both stand alone and MTL setting with contextual embeddings (BERT). All experiments use BiGRU.
}
\label{tab:srl-dyn-emb}
\end{table*}

\subsection{Contextual Embeddings}\label{sec:exp-contextual}
Table~\ref{tab:sprl-dyn-emb} summarizes our SPRL results using BERT contextual embeddings as input to the encoder RNN. Simply replacing the GloVe embeddings results in low performance on  \modelname{SPRL-only} 
(column~2). The model only predicts the majority class for the proto-roles with a severe class imbalance, and the F1 scores for those properties are 0. We observe that the model's weights are not updated during training as the loss on the validation set did not reduce. Hence most proto-roles show poor results.

We hypothesize that the issue is a model with too many parameters and relatively little training data. We therefore implement the model optimizations described in Section~\ref{sec:exp-description}. SPRL and SPR results for these experiments are shown in Tables~\ref{tab:sprl-dyn-emb} (Column 4, 5, 6) and Table~\ref{tab:srl-dyn-emb}, respectively. 
 In the multi-task setting, SPRL with gold heads and predicate representations from BERT alone performs worse than with static embeddings. However, adding sentence embeddings (here we simply use the [CLS] token representation from BERT; column~5) results in a big jump of over 10\% absolute, resulting in our best model and state-of-the-art at 87.3\% micro-averaged F1. For comparison, \citeposs{opitz-frank-2019-argument} ensemble setup, using gold SRL spans, achieves a micro-average F1-score of 86.8\% with their best model, slightly below our model that uses gold head information. 
 Using predicted head information, performance drops down to 83.9\% micro-F1.

We also briefly try removing the RNN encoder entirely in our SPRL-only baseline, relying on BERT's contextualized representations only. Here, we pass the concatenated encoding of the predicate and the argument head directly into a logistic regression unit for each proto-role. However, this model (Table~\ref{tab:sprl-dyn-emb}, column~3 `LR') underperforms both the GRU approach with BERT embeddings and the biLSTM approach with static embeddings. Future work might explore fine-tuning BERT for joint SRL and SPRL prediction specifically.

Table~\ref{tab:srl-dyn-emb} shows results for SRL with contextual embeddings and GRUs. Overall, the SRL-only results with gold spans/heads (columns~2-4) are similar to the static embedding results. We also see a similar drop with predicted spans and heads (columns~5-6), as well as a gain of about~1\% going from SRL-only to MTL (column~5 vs.\ 6).

Finally, we note that the GRU-based model is even faster to train than the biLSTM based model we used with static embeddings: it takes only 10 iteration to converge. Inference is fast too, owing to a neural network with relatively few parameters.

\section{Discussion}
Since we introduced  intermediate steps in our architecture, we examined their  prediction quality. 
The nameless span prediction was 89\% accurate, with 90\% macro-averaged F1-score.
The 
head prediction given gold argument spans had 99\% accuracy with 96\% macro-F1. On passing it the predicted argument spans, the head predictor was 98\% accurate with 93\% macro-F1. We noticed that adding predicted heads+spans, as opposed to gold ones, sometimes hurts role and proto-role prediction.

For efficiency and speed, we did not enforce the prediction of exactly one syntactic head per argument span. We analysed the head predictions and noticed only 27 out of 925 (2.9\%) head predictions were outside any argument span.
66 out of 862 spans had no head predicted and only 7 of them had 2 or more heads predicted (total 8\%).
This rate is higher than we expected and future work should improve on it, ideally without sacrificing speed.

Why are some of the MTL results for proto-roles with static word embeddings (using predicted span/heads)  lower than the replicated SPRL baseline?
One possible cause is the sparseness (class imbalance) of the "1"s in the binarized proto-role scores in training,  which may make learning harder, even more so with the MTL architecture which has more parameters than the baseline. For example, \rolename{changes\_possession} and \rolename{created} have only 75 and 65 instances of 1's (support), respectively, and their F1 decreased.
\rolename{stationary} (support 21) and \rolename{change\_of\_location} (support 52) do not show much improvement. But  proto-roles with the highest support, such as 
\rolename{existed\_before} (618), 
\rolename{existed\_during} (811), and
\rolename{existed\_after} (657),   
have the highest F1 gains.

\section{Conclusion}
We presented a  multi-task learning approach for SRL and SPRL, using either static  (GloVe)  or contextual  (BERT) embeddings.
We predicted intermediate nameless semantic argument spans and their syntactic heads to feed the joint task. 
We also used gold spans and heads to assess an upper bound for our approach.
We showed our approach's gains with both embeddings and achieved new state-of-the-art SPRL results 
after optimizing our network architecture for training with BERT on a rather small available dataset.
Moreover, unlike previous work, our model is independent of pre-training tasks and external SRL and/or syntactic parses as input, which makes it more widely applicable.

We saw that the intermediate steps as well as transfer learning within the steps in our pipeline help output quality, but not in all  architectures.

While our main focus 
was on improving SPRL, we showed that learning SPRL and SRL jointly, as well as using intermediate 
``nameless'' spans, improves SRL quality too. Explicit head information helps further, but only when using gold heads.  Future work can leverage our findings to improve on the state-of-the-art  of SRL.
We hope to further improve SPRL quality by using silver labels over large corpora and enforcing prediction of exactly one head per span without sacrificing speed.

\section*{Limitations}

While designing the model architecture, we used evaluation results on the test set to guide our decisions, even though specific hyper-parameters for each architecture were tuned on the separate dev set. This was due to the relatively small amount of available annotated data. While we believe prior work used a similar approach too, there is a risk that  architectural choices themselves could result in overfitting. Ideally our final model architecture presented in this paper should therefore be tested on an unseen test set, should additional appropriate labeled data become available.  

In our experiments, we used only data for which both SPRL and SRL annotations were available. Our model could in principle accommodate training inputs with only one type of annotation, by ignoring the loss from the component not present in a training instance. 
For example, training on
additional resources such as the UDS proto-role annotations \citep{white2016,white2020}, PropBank data not used by \citep{reisinger2015},
and/or the RW\_Eng corpus \citep{sayeed-etal-2018-rollenwechsel},
which contains large amounts of automatically annotated (`silver') SRL data.

Finally, 
our
approach  does not perform predicate identification. 
A model doing so will be even more applicable.

\section*{Ethics Statement}

With any semantic technology that can be used to extract information from raw text, it is important to emphasize that system performance is far from perfect. This kind of technology should not be used, without human review, in situations where it has an immediate effect on decision making. As in any machine learning approach, there is a potential for biases in the training data to be reflected in model predictions. This problem is amplified when using crowdsourced annotations (such as the proto-role annotations in the data we use), because quality control and documentation is more challenging. Another potential issue is that the data used in this work is ultimately based on the Wall Street Journal, which is reflective only of a particular domain, genre, and register, representing a narrow slice of English speakers. As a result, our models are not immediately applicable to other types of input, at least not with an expectation of similar performance. We also did not run experiments on languages other than English. 
We do however, believe that the methods presented in this paper will generalize to other types of training data and even other languages, in particular because our end-to-end model makes few assumptions about the availability of syntactically annotated inputs.  The work on proto-roles, in general, has the potential to result in semantic annotations that are more theory agnostic and language independent, potentially avoiding biases inherent in specific semantic theories.


\bibliography{anthology,custom}
\bibliographystyle{acl_natbib}

\appendix



\end{document}